\documentclass[letterpaper]{article} 
\usepackage{aaai25}  
\usepackage{times}  
\usepackage{helvet}  
\usepackage{courier}  
\usepackage[hyphens]{url}  
\usepackage{graphicx} 
\usepackage{natbib}  
\usepackage{caption} 
\usepackage{algorithm}
\usepackage{algorithmic}
\usepackage{times}
\usepackage{helvet}
\usepackage{courier}
\usepackage{microtype}
\usepackage{amsfonts}
\usepackage{amssymb}
\usepackage{multirow}
\usepackage{graphicx}
\usepackage{amssymb}
\usepackage{natbib}
\usepackage{amsmath}
\usepackage{appendix}
\usepackage{subfigure}
\usepackage{booktabs}
\usepackage{listings}
\usepackage{amsmath}
\usepackage[table]{xcolor}
\usepackage{newfloat}
\usepackage{listings}
\DeclareCaptionStyle{ruled}{labelfont=normalfont,labelsep=colon,strut=off} 
\floatstyle{ruled}
\newfloat{listing}{tb}{lst}{}
\floatname{listing}{Listing}
\title{Unleashing the Potential of Model Bias for Generalized Category Discovery}
\author{
    Wenbin An\textsuperscript{\rm 1,2}\footnotemark[1], Haonan Lin\textsuperscript{\rm 1,2}\thanks{Equal contribution.}, Jiahao Nie\textsuperscript{\rm 3},
    Feng Tian\textsuperscript{\rm 1,2}\footnotemark[2], Wenkai Shi\textsuperscript{\rm 1,2}, \\Yaqiang Wu\textsuperscript{\rm 4}, Qianying Wang\textsuperscript{\rm 4}\thanks{Corresponding authors.}, Ping Chen\textsuperscript{\rm 5}\\
}
\affiliations{

    \textsuperscript{\rm 1}Xi'an Jiaotong University 
    \textsuperscript{\rm 2}National Engineering Laboratory for Big Data Analytics \\
    \textsuperscript{\rm 3}Nanyang Technological University
    \textsuperscript{\rm 4}Lenovo Research 
    \textsuperscript{\rm 5}University of Massachusetts Boston\\
    \{wenbinan,linhaonan,kkkkkkai0611\}@stu.xjtu.edu.cn, jiahao007@e.ntu.edu.sg, \\ fengtian@mail.xjtu.edu.cn, 
    \{wuyqe,wangqya\}@lenovo.com, Ping.Chen@umb.edu
}

\usepackage{bibentry}

\begin{document}

\maketitle

\begin{abstract}
Generalized Category Discovery is a significant and complex task that aims to identify both known and undefined novel categories from a set of unlabeled data, leveraging another labeled dataset containing only known categories. 
The primary challenges stem from model bias induced by pre-training on only known categories and the lack of precise supervision for novel ones, leading to \textit{category bias} towards known categories and \textit{category confusion} among different novel categories, which hinders models' ability to identify novel categories effectively.
To address these challenges, we propose a novel framework named \textit{Self-Debiasing Calibration} (SDC). Unlike prior methods that regard model bias towards known categories as an obstacle to novel category identification, SDC provides a novel insight into unleashing the potential of the bias to facilitate novel category learning. 
Specifically, we utilize the biased pre-trained model to guide the subsequent learning process on unlabeled data. 
The output of the biased model serves two key purposes. First, it provides an accurate modeling of category bias, which can be utilized to measure the degree of bias and debias the output of the current training model.
Second, it offers valuable insights for distinguishing different novel categories by transferring knowledge between similar categories.
Based on these insights, SDC dynamically adjusts the output logits of the current training model using the output of the biased model. 
This approach produces less biased logits to effectively address the issue of category bias towards known categories, and generates more accurate pseudo labels for unlabeled data, thereby mitigating category confusion for novel categories.
Experiments on three benchmark datasets show that SDC outperforms SOTA methods, especially in the identification of novel categories.
Our code and data are available at \url{https://github.com/Lackel/SDC}.
\end{abstract}

\begin{figure}[t]
\centering
\includegraphics[width=0.45\textwidth]{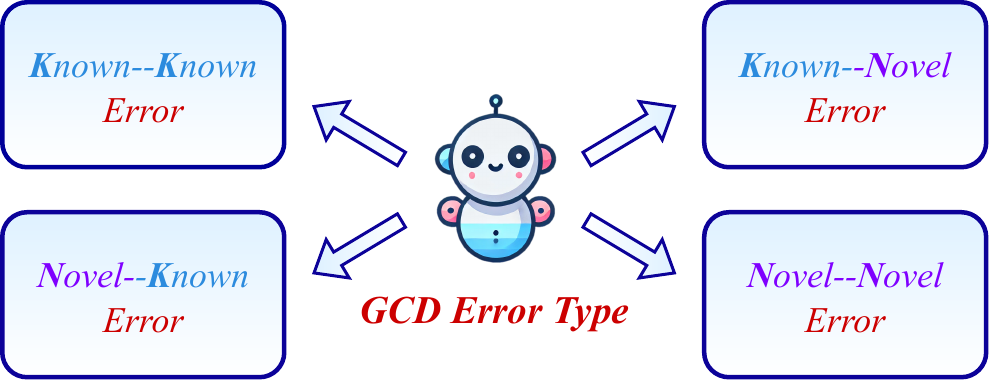}
\captionof{figure}{Four types of GCD errors. For instance, \textit{Novel-Novel Error} indicates samples from novel categories are misclassified into wrong novel categories.}
\label{fig:cm}
\end{figure}

\section{Introduction}
Unlike humans who can transfer previously acquired knowledge to new concepts, most deep learning models fail to recognize new categories due to their closed-world assumption that training and test data come from the same predefined categories. The assumption limits their applications in numerous practical tasks such as intent detection \citep{diffusion,down} and image categorization \citep{vaze2024no,rastegar2024learn}, where well-trained models may encounter unlabeled data from previously unseen novel categories.
To address the requirements of open-world scenarios, Generalized Category Discovery (GCD) has been proposed and extensively studied in both NLP \citep{dpn,zhang2024ronid} and CV fields \citep{zhao2024pseudo,tribranch,flip}.
GCD aims to recognize both known and novel categories from unlabeled data, using only a few labeled samples from known categories. This task enables models to adapt to an increasing number of novel categories without incurring additional labeling costs.

To address GCD, existing research \cite{ncl,ktn,tan} typically begins with pre-training models on labeled data tailored to specific tasks. However, in the absence of labeled data for novel categories, the pre-trained models exhibit a strong bias towards known ones, resulting in the misclassification of samples from novel categories into known ones. To alleviate this issue, existing work fine-tunes the pre-trained models on unlabeled data through pseudo-label training \cite{thu2021}, contrastive learning \cite{gcd,dpn}, and transfer learning \cite{ktn,tan}. Despite performance improvements, these models still struggle with identifying novel categories due to inherent model bias towards known categories and the lack of supervision for novel ones.

In this paper, we first provide a novel insight into the model bias towards known categories: \textit{Can we leverage this bias to enhance model performance on novel categories, rather than simply mitigate it like previous methods?} This insight may seem counter-intuitive since model bias is generally perceived as detrimental to novel categories. To explore this question, we first delve into the source of errors in GCD and mainly focus on errors for novel categories. As shown in Fig. \ref{fig:cm}, the error for novel categories can be divided into two parts: samples misclassified into known categories and samples misclassified into other novel categories. The former primarily arises from \textbf{category bias} towards known categories induced by pre-training \cite{mtp}, while the latter is largely due to \textbf{category confusion} among novel categories resulting from the absence of supervision.

To mitigate the issues of category bias and category confusion, we propose a novel framework named \textit{Self-Debiasing Calibration} (SDC) with the insight of unleashing the potential of model bias to enhance model performance on novel categories. 
Specifically, SDC first investigates the question: \textit{What can we learn from the biased model pre-trained on labeled data for subsequent training on unlabeled data?}
On the one hand, the output of the biased model \textbf{provides an accurate representation of model bias}, which can precisely measure the degree of bias and be utilized to debias the output of the current training model, thereby recovering less biased predictions for unlabeled data. This insight suggests that category bias itself can be harnessed to mitigate the issue of category bias.
On the other hand, the output of the biased model \textbf{offers insightful hints for distinguishing between different novel categories} by transferring knowledge between similar categories. For example, given the known categories `dog' and `goose', and the novel categories `cat' and `duck', if the biased model predicts a sample as `dog', then the sample is more likely to be a `cat' than a `duck' due to the greater similarity between cats and dogs. This example shows that predictions from the biased model can help mitigate the issue of category confusion among novel categories.
Based on the two insights, SDC dynamically adjusts the output logits of the current training model using the output of the biased model, which can help to produce less biased logits and more accurate pseudo labels for unlabeled data, thereby effectively addressing the issues of category bias and category confusion. 
To avoid punishing instances from known categories in unlabeled data, we employ an entropy-based weighting mechanism \cite{ktn} to effectively distinguish between known and novel categories.
Furthermore, thanks to the classifier-based architecture, SDC can perform online inference with greater convenience and lower inference latency compared to previous clustering-based methods \cite{tan}. 
Experiments conducted on three benchmark datasets demonstrate that SDC outperforms SOTA methods, particularly in the accurate identification of novel categories.

Our main contributions can be summarized as follows:
\begin{itemize}
  \item We provide a novel insight into unleashing the potential of model bias to mitigate category bias and category confusion for novel categories in GCD. 
  \item We introduce \textit{Self-Debiasing Calibration} (SDC), which incorporates two logit adjustment techniques to effectively leverage model bias for learning novel categories.
  \item Experiments demonstrate that our model outperforms SOTA methods, especially in identifying novel categories.
\end{itemize}

\section{Related Work}
\subsection{Generalized Category Discovery}
Generalized Category Discovery (GCD) is a critical and challenging task in the open-world setting, which posits that the encountered unlabeled data contain unseen novel categories \cite{gcd,dpn} or fine-grained categories \cite{fcdc,dna}. Previous models have predominantly utilized pre-training on labeled data for model initialization. Nevertheless, the pre-trained model tends to be biased towards known categories due to the absence of supervision for novel ones \cite{tan}. 
To address model bias and enhance the acquisition of knowledge about novel categories from unlabeled data, prior models have incorporated pseudo-label training \cite{dec,thu2021}, contrastive learning \cite{gcd,dpn}, and transfer learning \cite{tan,zhang2024new} methods.
For instance, \citet{loop} implemented neighborhood contrastive learning with the help of LLMs to improve representation learning for unlabeled data. \citet{ktn} proposed transferring knowledge from known to novel categories through label adjustment, while \citet{tan} employed category prototypes to measure category similarities and facilitate knowledge transfer.
Despite the improved performance on known categories, these methods exhibit sub-optimal performance on novel categories due to persistent model bias and the lack of supervision.

\subsection{Logit Adjustment}
Logit adjustment seeks to mitigate issues related to model bias towards specific output by modifying the logits. This technique is widely employed in scenarios involving long-tailed learning \cite{tail1} and zero-shot learning \cite{zero1}. For example, \citet{tail4} proposed a class-balanced loss based on the effective number of samples, where logit adjustment is applied to counteract the class bias. \citet{tail2} developed a novel adaptive logit adjustment loss that considers both the quantity and difficulty of data. In the context of zero-shot learning, \citet{zero1} exploited semantic priors in logit adjustment to acquire knowledge about unseen categories. Additionally, \citet{zero2} introduced a dynamic few-shot learning approach where the logits of novel categories are adjusted to enhance performance on these categories. For GCD, \citet{ktn} proposed transferring logits and labels from known to novel categories to mitigate prediction bias. 
Despite the improved performance, these methods struggle to reconstruct unbiased logits relying only on the original ones.

\begin{figure*}
\centering
\includegraphics[width=0.9\textwidth]{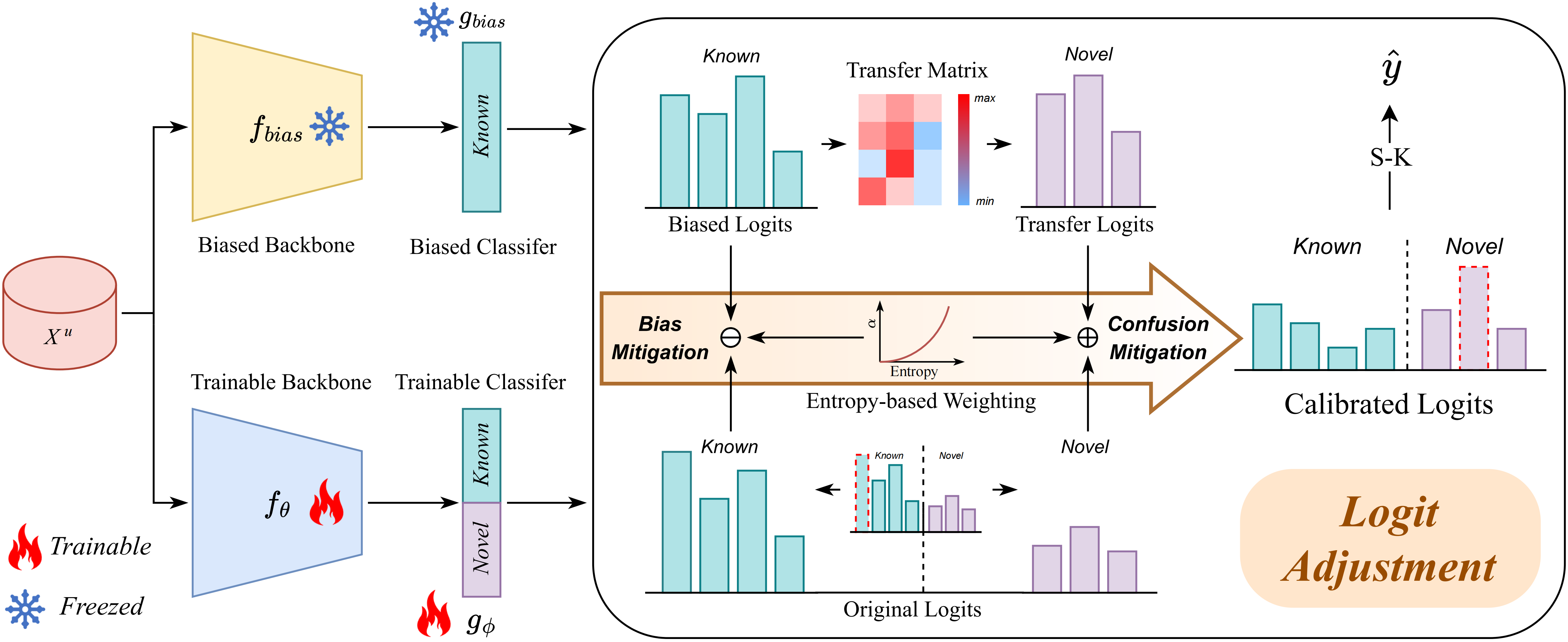}
\caption{The overall framework of our model, which mainly contains Bias Mitigation, Confusion Mitigation, and Entropy-based Weighting. $\ominus$ means subtracting the biased logits from the original ones for known categories, while $\oplus$ means adding the transfer logits to the original ones for novel categories. $\hat{y}$ is the generated pseudo label utilizing the calibrated logits through the Sinkhorn-Knopp (S-K) algorithm.} 
\label{fig2}
\end{figure*}

\section{Method}
\subsubsection{Problem Formulation.}
Models trained on a labeled dataset $\mathcal{D}^{l} = \{(x_{i},y_{i}) \mid y_{i} \in \mathcal{Y}_{k}\}$ are proficient at identifying predefined known categories $\mathcal{Y}_{k}$. However, in real-world scenarios,  these models often encounter unlabeled data $\mathcal{D}^{u} = \{x_{i} \mid y_{i} \in {\mathcal{Y}_{k} \cup \mathcal{Y}_{n}}\}$ that encompasses both known categories $\mathcal{Y}_{k}$ and novel categories $\mathcal{Y}_{n}$ ($\mathcal{Y}_{k} \cap \mathcal{Y}_{n} = \emptyset$),  potentially leading to identification failure. 
To overcome this challenge, Generalized Category Discovery (GCD) requires models to recognize both known and novel categories based on $\mathcal{D}^{l}$ and $\mathcal{D}^{u}$, despite the absence of annotations for the novel categories. 
We denote the number of known categories as $M = |\mathcal{Y}_{k}|$, the number of novel categories as $N = |\mathcal{Y}_{n}|$, and the total number of categories as $K = |\mathcal{Y}_{k} \cup \mathcal{Y}_{n}|$. We assume $K$ is known following previous work \cite{dpn,ktn} and solve the problem of $K$ estimation in the later section. Finally, the model's performance will be evaluated on a testing set $\mathcal{D}^{t} = \{(x_{i},y_{i}) \mid y_{i} \in {\mathcal{Y}_{k} \cup \mathcal{Y}_{n}}\}$ in an inductive manner.

\subsubsection{Framework Overview.} 
To address category bias and confusion for GCD, we design \textit{Self-Debiasing Calibration} (SDC), depicted in Fig. \ref{fig2}. Initially, we conduct pre-training on labeled data to obtain a biased backbone $f_{bias}$ and a biased classifier $g_{bias}$. Subsequently, we initialize a trainable backbone $f_{\theta}$ with a trainable classifier $g_{\phi}$ to learn from unlabeled data. 
We then refine the original logits from the trainable classifier by utilizing logits from the biased classifier, thereby producing less biased logits.
To mitigate category bias, we subtract the biased logits from the original ones for known categories. Conversely, for novel categories, we add transfer logits to the original ones, where the transfer logits are calculated via the biased logits and a transfer matrix grounded in category similarities. 
Additionally, to prevent penalizing samples from known categories in unlabeled data, we implement a weighting mechanism based on the output entropy of the biased classifier. 
Following logit adjustment, we generate pseudo labels for unlabeled data utilizing the calibrated logits through the Sinkhorn-Knopp (S-K) algorithm.

\subsection{Model pre-training and Initialization}
\subsubsection{Model pre-training.}
We employ the pre-trained Bert \citep{bert} as our backbone and utilize a linear layer as the classifier. To adapt the pre-trained model for the downstream GCD task, we apply cross-entropy (CE) loss on labeled data and masked language modeling (MLM) loss \citep{pretrain} on unlabeled data for pre-training. The CE loss facilitates the learning of domain-specific knowledge, while the MLM loss aids in acquiring general knowledge. Given that the labeled data encompasses only known categories, the pre-trained model and classifier can exhibit significant bias towards these categories \cite{ktn,tan}. Therefore, we denote them as the biased backbone $f_{bias}: \mathcal{X} \rightarrow \mathbb{R}^{d}$ and the biased classifier $g_{bias}: \mathbb{R}^{d} \rightarrow \mathbb{R}^{M}$, where $d$ represents the feature dimension and $M$ is the number of known categories.

\subsubsection{Model Initialization.}
After pre-training, we freeze the parameters of $f_{bias}$ and $g_{bias}$ to guide subsequent training. We then employ another trainable backbone $f_{\theta}: \mathcal{X} \rightarrow \mathbb{R}^{d}$, along with a trainable classifier $g_{\phi}: \mathbb{R}^{d} \rightarrow \mathbb{R}^{K}$ for model training, where $\theta$ and $\phi$ are learnable parameters, and $K$ represents the total number of categories. The parameters of $f_{\theta}$ are initialized using those of $f_{bias}$. Given the absence of parameters for novel categories, we initialize the parameters of $g_{\phi}$ with category prototypes. 
Specifically, we perform KMeans clustering on the unlabeled data and treat the cluster centers as prototypes. To align the indices of known categories in $g_{bias}$ and $g_{\phi}$, we employ an alignment algorithm following \citet{dpn}. Subsequently, $g_{\phi}$ is initialized with the $\ell_2$-normalized  aligned prototypes:  $g_{\phi} = [\mathbf{c_{1}}, ..., \mathbf{c_{M}}, ..., \mathbf{c_{K}}]$, where the indices from 1 to $M$ correspond to the prototypes of known categories and the indices from $M+1$ to $K$ correspond to the prototypes of novel categories.

\subsection{Self-Debiasing through Logit Adjustment}
Without fine-tuning for novel categories, $f_{\theta}$ and $g_{\phi}$ can exhibit significant bias towards known categories. This bias can be exacerbated if the output of the biased model is used for training, thus necessitating the adjustment of the biased output to provide more accurate supervision for unlabeled data.
In this section, we introduce two methods to calibrate the original logits based on the biased logits, aimed at mitigating category bias and category confusion, respectively.
To prevent penalizing samples from known categories, we employ an entropy-based weighting mechanism, which effectively distinguishes samples from known and novel categories. 

\subsubsection{Category Bias Mitigation.}
Models pre-trained on labeled data with only known categories often exhibit bias and overconfidence toward these categories \cite{tan}. As a result, samples from novel categories within unlabeled data are frequently misclassified into known categories \cite{ktn}. This issue, referred to as \textbf{category bias}, can significantly impede the model's ability to learn novel categories from unlabeled data and adversely affect its performance on these categories. A simple approach to mitigate category bias involves subtracting a penalty factor from the logits of all known categories. However, this indiscriminate penalization does not account for the varying degrees of bias among different known categories, potentially leading to sub-optimal outcomes with either insufficient or excessive penalties. 
To address these limitations, we propose to leverage the output logits of the biased model, which can accurately measure the degree of category bias and offer a precise representation of category bias. 
By contrasting the logits of the current model with those of the biased model, our approach can facilitate the generation of less biased predictions and more accurate pseudo-labels for model training.
Specifically, given the logits $\mathbf{L_{\theta}}$ from the current model and the biased logits $\mathbf{L_{bias}}$ from the biased model, the logits after category bias mitigation can be expressed as follows:

\begin{equation}
    \mathbf{L_{\theta}} = g_{\phi}(f_{\theta}(x))
\end{equation}

\begin{equation}
    \mathbf{L_{bias}} = g_{bias}(f_{bias}(x))
\end{equation}

\begin{equation}
    \mathbf{L_{CBM}} = \mathbf{L_{\theta}} \left[:M \right] - \alpha \cdot \mathbf{L_{bias}}
\end{equation}
where $f_{\theta}$ and $g_{\phi}$ represent the current trainable backbone and classifier, $f_{bias}$ and $g_{bias}$ denote the frozen biased backbone and classifier, $x$ is the input instance, and $\alpha$ is a weighting factor. $\mathbf{L_{\theta}} \left[:M \right]$ means the first $M$ elements of $\mathbf{L_{\theta}}$ that correspond to known categories.

\subsubsection{Category Confusion Mitigation.}
Although Category Bias Mitigation (CBM) can reduce model bias towards known categories, distinguishing between different novel categories remains challenging due to the lack of supervision, which results in close logits for these categories. This issue, referred to as \textbf{category confusion}, can easily lead models to misclassify instances from novel categories into wrong novel categories.  
To address this issue, we propose to transfer knowledge from known categories to novel ones with the help of the biased model. The outputs of the biased model provide valuable insights for differentiating novel categories by considering similarities between known and novel categories. For instance, an instance from novel categories is more likely to be a `cat' rather than a `duck' if the biased model predicts it as a `dog', given the higher similarity between cats and dogs. Specifically, we use prototypes to measure category similarities, and the transfer matrix $\mathbf{T}$ is calculated as follows:

\begin{equation}
    \mathbf{T}_{M \times N} = \begin{bmatrix}
                    \mathbf{c_{1}}^\top \mathbf{c_{M+1}} & \mathbf{c_{1}}^\top \mathbf{c_{M+2}} & \cdots & \mathbf{c_{1}}^\top \mathbf{c_{M+N}} \\
                    \mathbf{c_{2}}^\top \mathbf{c_{M+1}} & \mathbf{c_{2}}^\top \mathbf{c_{M+2}} & \cdots & \mathbf{c_{2}}^\top  \mathbf{c_{M+N}} \\
                    \vdots & \vdots & \ddots & \vdots \\
                    \mathbf{c_{M}}^\top \mathbf{c_{M+1}} & \mathbf{c_{M}}^\top  \mathbf{c_{M+2}} & \cdots & \mathbf{c_{M}}^\top  \mathbf{c_{M+N}} \\
\end{bmatrix}
\end{equation}
where $\{\mathbf{c_{1}}, ..., \mathbf{c_{M}}\}$ and $\{\mathbf{c_{M+1}}, ..., \mathbf{c_{M+N}}\}$ are prototypes for known and novel categories, respectively. $M$ and $N$ denote the number of known and novel categories, respectively. We further normalize $\mathbf{T}$ along each row to obtain the transfer probabilities of each known category to different novel categories.
The logits after Category Confusion Mitigation can be expressed as follows:
\begin{equation}
    \mathbf{L_{CCM}} = \mathbf{L_{\theta}} \left[M : \right] + \alpha \cdot \mathbf{T^\top} \times \mathbf{L_{bias}}
\end{equation}
where $\mathbf{L_{\theta}} \left[M: \right]$ means the last $N$ elements of $\mathbf{L_{\theta}}$ that correspond to novel categories.
Combining category bias mitigation and category confusion mitigation, the final calibrated logits can be obtained by concatenating the two logits:
\begin{equation}
    \mathbf{L_{c}} = \left[ \mathbf{L_{CBM}} ,  \mathbf{L_{CCM}}  \right]
\end{equation}

\begin{figure}[t]
\centering
\includegraphics[width=0.35\textwidth]{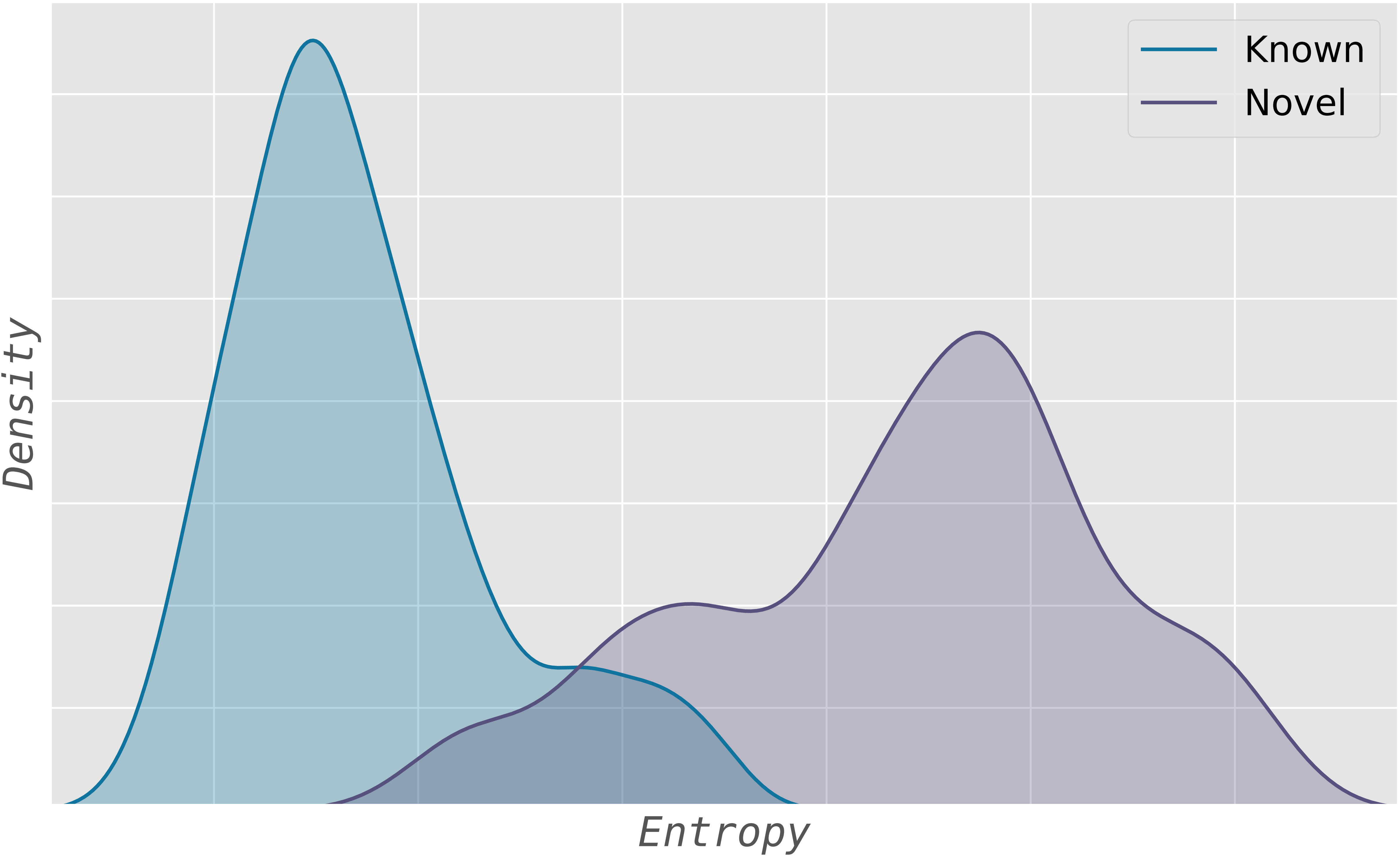}
\captionof{figure}{Entropy distribution for known and Novel categories based on the biased logits.}
\label{fig3}
\end{figure}

\begin{table*}
\centering
\tabcolsep=0.3cm
\begin{tabular}{lccccccccc}

\toprule
\multirow{2}*{Method} & \multicolumn{3}{c}{BANKING} &\multicolumn{3}{c}{HWU64} & \multicolumn{3}{c}{CLINC}\\ 
\cmidrule(r){2-4}  \cmidrule(r){5-7}  \cmidrule(r){8-10}
            &H-score    &Known    &Novel    &H-score    &Known    &Novel    &H-score    &Known    &Novel \\
\midrule
DeepCluster &13.97  &13.94  &13.99  &15.27  &17.04  &13.84  
&26.48  &27.34  &25.67  \\
DCN         &16.33  &18.94  &18.78  &29.22  &21.63  &16.59  
&29.20  &30.00  &28.45  \\
DEC         &17.82  &20.36  &15.84  &19.11  &23.47  &16.11  
&19.78  &20.18  &19.40  \\
BERT     &21.08  &21.48  &20.70  &32.00  &36.65  &28.40  
&34.05  &34.98  &33.16  \\
GloVe    &29.25  &29.11  &29.39  &35.75  &35.08  &36.44  
&51.62  &51.74  &51.50  \\
SAE         &37.77  &38.29  &37.27  &39.47  &44.53  &35.45  
&45.74  &47.35  &44.24  \\
\midrule
GPT-3.5  &17.62  &62.24  &10.26  &28.30  &72.39  &17.59  &26.64  &61.29 &17.02  \\
GPT-4  &26.20  &80.13  &15.66  &33.80  &80.39  &21.40  &26.39  &90.77 &15.44  \\

\midrule
Simple   &40.52  &49.96  &34.08  &-  &-  &-  
&62.76  &70.60  &56.49  \\
Semi-DC  &47.40  &53.37  &42.63  &51.24  &54.41  &48.42  
&73.41  &75.60  &71.34  \\
CDAC+    &50.28  &55.42  &46.01  &55.37  &58.14  &52.86  
&69.42  &70.08  &68.77  \\
DTC      &52.13  &59.98  &46.10  &59.80  &64.94  &55.42  
&68.71  &82.34  &58.95  \\
Semi-KM  &54.83  &73.62  &43.68  &69.11  &80.30  &60.66  
&70.98  &89.03  &59.01  \\
DAC      &54.98  &69.60  &45.44  &64.79  &77.55  &55.64  
&78.77  &89.10  &70.59  \\
GCD      &55.78  &75.16  &44.34  &67.28  &78.58  &58.82  
&63.08  &89.64  &48.66  \\
PTJN     &60.69  &77.20  &50.00  &-  &-  &-  
&83.34  &91.79  &76.32  \\
DPN      &60.73  &80.93  &48.60  &72.55  &79.77  &66.52  
&84.56  &92.97  &77.54  \\
TAN     &66.70  &81.97  &56.23  &70.68  &80.68  &62.89  &87.02  &93.39 &81.46  \\
KTN     &68.20  &81.65  &58.55  &76.34  &80.15  &72.88  &86.13  &93.93 &79.53  \\

\midrule
\textbf{SDC} (\textbf{Ours})        &\textbf{70.11}  &\textbf{82.16}  &\textbf{61.14}  &\textbf{79.81}  &\textbf{82.08}  &\textbf{77.66}  &\textbf{87.65}  &\textbf{94.12}  &\textbf{82.02}  \\
 
\bottomrule
\end{tabular}
\caption{Average results (\%) over 3 runs on the testing sets. The best performance is highlighted in bold.}
\label{table1}
\vspace{-3mm}
\end{table*}

\subsection{Entropy-based Weighting}
Given that the unlabeled dataset contains both known and novel categories, the aforementioned logit adjustment methods might penalize instances from known categories, thus degrading model performance on these categories. Therefore, effectively distinguishing between known and novel categories in unlabeled data and setting the adjustment weight $\alpha$ appropriately is crucial for balancing model performance across different categories. Inspired by the findings of \citet{ktn}, we use the entropy of the biased logits as an indicator to differentiate samples from known or novel categories and decide the adjustment weights. As shown in Fig. \ref{fig3}, samples from known categories exhibit low entropy because the biased model is confident in these categories. Conversely, samples from novel categories exhibit high entropy since the biased model has not encountered these types of samples before, resulting in relatively uniform logits.
Based on these insights, we employ the entropy of each instance as the weighting factor $\alpha$ for CBM and CCM. 
The entropy $E_{i}$ for each instance $x_{i}$ is calculated as follows:
\begin{equation}
    E_{i} = -\sum_{j} p_{ij} \log p_{ij}
\end{equation}
where $p_{i} = \textrm{Softmax} (\mathbf{L}_{bias}(x_{i}))$ is the probability distribution for instance $x_{i}$ from the biased model. The weighting factor $\alpha_{i}$ for instance $x_{i}$ is then defined as:

\begin{equation}
    \alpha_{i} = \beta \cdot \textrm{Sigmoid} (E_{i} - E_{max})
\end{equation}
where $\beta$ is a fixed weighting factor, and $E_{max}$ is the maximum entropy in the current batch of data. The Sigmoid function helps in scaling the entropy difference.

\subsection{Discussion}
Different from previous work that treats model bias towards known categories as detrimental to novel categories, we explore the potential of the biased model for learning about novel categories from three aspects. 
First, the output of the biased model provides an accurate representation of category bias, which can precisely measure the degree of bias toward different categories and be utilized to recover unbiased outputs from the current model.
Second, by transferring knowledge between similar categories, the output of the biased model can offer insightful hints for distinguishing between different novel categories.
Third, the entropy of instances predicted by the biased model is an effective indicator for distinguishing between known and novel categories, which helps to avoid punishing instances from known categories.
Based on these insights, our \textit{Self-Debiasing Calibration} (SDC) develops two logit adjustment methods with a weighting mechanism to mitigate the issues of category bias and category confusion, significantly enhancing model performance on novel categories.

\subsection{Pseudo-label Generation and Model Training}
Given the calibrated logits $\mathbf{L_{c}}$, pseudo labels $\hat{Y}$ for unlabeled data are generated by solving an optimal transport problem \cite{pseudo}, which promotes a uniform distribution of pseudo labels across all $K$ categories, thereby mitigating category bias:

\begin{equation}
\begin{gathered}
\hat{\mathbf{Y}} = \max_{\mathbf{Y}} \text{Tr}(\mathbf{Y}\mathbf{L_{c}}) + \epsilon \cdot \textrm{E}(\mathbf{Y}) \\
\text{s.t.} \quad \mathbf{Y} \in \mathbb{R}_{+}^{B \times K}, \mathbf{Y}^\top \mathbf{1}_B = \frac{1}{K}\mathbf{1}_{K}, \mathbf{Y} \mathbf{1}_{K} = \frac{1}{B}\mathbf{1}_B
\end{gathered}
\label{eq9}
\end{equation}
where Tr denotes the trace of the matrix, $\epsilon$ is a weighting factor, and $\textrm{E}$ represents the entropy function. $B$ and $K$ indicate the batch size and the number of categories, respectively. The vectors $\mathbf{1}_B$ and $\mathbf{1}_K$ are unit vectors of dimension $B$ and $K$, respectively. The solution of Eq. \ref{eq9} can be derived using the Sinkhorn-Knopp algorithm \cite{sk}.
Subsequently, we apply cross-entropy loss to both labeled and unlabeled data for representation learning:
\begin{equation}
    \mathcal{\ell}_{sup} = \lambda_{1} \cdot \mathcal{\ell}_{ce}(\hat{Y}^{u}, \hat{Y}) + (1 - \lambda_{1}) \cdot \mathcal{\ell}_{ce}(\hat{Y}^{l}, Y^{l})
\end{equation}
where $\lambda_{1}$ is a factor that balances the weights between labeled and unlabeled data, $ \hat{Y}^{l}$ and $\hat{Y}^{u}$ represent predictions based on the original logits for labeled and unlabeled data, respectively. $Y^{l}$ denotes ground-truth labels for labeled data. We also employ instance-level contrastive learning $\ell_{cont}$ on both labeled and unlabeled data to learn more consistent representations. The overall training objective of our model is formalized as:
\begin{equation}
    \mathcal{\ell } = \mathcal{\ell }_{sup} + \lambda_{2} \cdot \mathcal{\ell }_{cont}
\end{equation}
where $\lambda_{2}$ is a weighting factor for the contrastive loss. 

\section{Experiments}
\subsection{Experimental Setup}
\subsubsection{Datasets.}
We perform experiments on three generic text categorization datasets: \textbf{BANKING} is a fine-grained intent detection dataset in the banking domain \cite{banking}. \textbf{HWU64} is an assistant query classification dataset \cite{hwu64}. \textbf{CLINC} is a multi-domain intent detection dataset \citep{clinc}. We randomly select 25\% categories as novel categories and 10\% data as labeled data following previous work \cite{thu2021,dpn}. 

\subsubsection{Comparison with SOTA models.}
We compare our model with various baselines and SOTA methods.

\noindent \textbf{Unsupervised Models.}\quad  DeepCluster \citep{deepcluster}; DCN \citep{dcn}; DEC \citep{dec}; BERT \citep{bert}; GloVe \citep{glove}; SAE \cite{thu2021}.

\noindent \textbf{Semi-supervised Models.}\quad Simple \cite{simple}; Semi-DC \cite{deepcluster}; CDAC+ \cite{thu2020}; DTC \cite{dtc}; Semi-KM \cite{km}; DAC \citep{thu2021}; GCD \citep{gcd}; PTJN \cite{ptjn}; DPN \citep{dpn}; TAN \cite{tan}; KTN \cite{ktn}.

\noindent \textbf{GPT-based Models.}\quad GPT-3.5 \cite{gpt3}; GPT-4 \cite{gpt4}.

\subsubsection{Evaluation Metrics.}
We evaluate the model performance with clustering accuracy following standard practice \citep{dpn,ktn}.
(1) \textbf{Known}: clustering accuracy for known categories.
(2) \textbf{Novel}: clustering accuracy for novel categories.
(3) \textbf{H-score}: harmonic mean of the clustering accuracy for known and novel categories to avoid biased evaluation towards known ones \citep{hscore}.

\subsubsection{Implementation Details.}
We use the pre-trained Bert-base-uncased model \citep{huggingface} as our backbone. We fine-tune the last four Transformer layers and the classifier layer using the AdamW optimizer. Early stopping is employed during pre-training with a patience of 20 epochs.
For hyper-parameters, $\beta$ is set to 0.03, 0.05, and 0.42 for the BANKING, HWU64, and CLINC datasets, respectively. $\lambda_{1}$ is set to 0.6 and gradually increases to 0.7, $\lambda_{2}$ is set to 0.01. Following previous work \cite{dpn,ktn}, the pre-training and training epochs are set to 100 and 80, respectively. The learning rate for both pre-training and training is set to $5e^{-5}$. The batch size is set to 128 for pre-training and training. For the implementation of the Sinkhorn-Knopp algorithm, we follow the settings in \citet{sk2}. GPT-3.5 and GPT-4 are based on the turbo API of OpenAI.
All the experiments are conducted on a single RTX-3090 GPU.

\begin{table}[t]
\label{tab:gpt4}
\centering
\begin{tabular}{lccc}
\toprule
Variant                & H-score & Known & Novel  
\\ \midrule
\rowcolor{gray!20} SDC & 79.81 & 82.08  & 77.66 \\
\midrule
\multicolumn{4}{c}{\textit{\textbf{pre-training}}} \\
w/o MLM                & 74.73 & 79.58 & 70.43  \\
w/o pre-training       & 68.12 & 78.48 & 60.18  \\
w/o CE                 & 64.09 & 78.56 & 54.12  \\
\midrule
\multicolumn{4}{c}{\textit{\textbf{Logit Adjustment (LA)}}} \\
w/o CCM               & 78.40 & 80.70 & 76.23  \\
w/o CBM               & 77.78 & 81.91 & 74.05  \\
w/o Weighting         & 77.43 & 76.90 & 77.97  \\
w/o LA                & 76.79 & 81.16 & 72.87  \\
\midrule
\multicolumn{4}{c}{\textit{\textbf{Training}}} \\
w/o $\mathcal{\ell}_{cont}$        & 78.99 & 81.90 & 76.28 \\
w/o Initialization                 & 76.78 & 82.74 & 71.62  \\
w/o $\mathcal{\ell}_{ce}(X^{l})$   & 75.02 & 77.33 & 72.84 \\
w/o $\mathcal{\ell}_{ce}(X^{u})$   & 58.44 & 83.73 & 44.88  \\
\midrule
\multicolumn{4}{c}{\textit{\textbf{Real-world Applications}}} \\
w/o Ground-truth $K$   & 79.34 & 82.11 & 76.75  \\
Online Inference   & 78.94 & 83.08 & 75.20  \\
\bottomrule
\end{tabular}
\caption{Results of ablation studies on the HWU64 dataset.}
\label{table2}
\end{table}

\subsection{Results and Analysis}
\subsubsection{Comparison with the SOTA models.}
We present the performance of different models in Table \ref{table1}. Our model demonstrates superior performance across all datasets and evaluation metrics. Notably, there is an average improvement of 2.64\% in the accuracy of novel categories compared to SOTA methods, highlighting the efficacy of our logit adjustment framework. Furthermore, our model shows an average improvement of 0.28\% in the accuracy of known categories, which indicates that the weighting strategy employed by our model enhances performance on novel categories without compromising the accuracy of known categories. Consequently, our model achieves an average 2.00\% improvement in the H-score, indicating a balanced recognition capability for both known and novel categories. Furthermore, GCD models in the CV fields (i.e., Simple \cite{simple} and GCD \cite{gcd}) perform poorly on the textual data, which motivates us to develop models tailored for the NLP domain.

\subsubsection{Comparison with ChatGPT.}
Given the impressive zero-shot capabilities of large language models (LLMs), we include GPT-3.5 and GPT-4 as baselines for comparison. Despite the competitive performance on known categories, LLMs face challenges in recognizing novel categories due to the absence of prior knowledge. This observation underscores the importance of accurately identifying novel categories and further demonstrates the effectiveness of our model.

\subsubsection{Ablation Study.}
We conduct ablation studies from four perspectives in Table \ref{table2} to validate the effectiveness of each component in our model.
\textbf{Pre-training} plays a crucial role in model initialization, providing biased logits that facilitate learning about novel categories.
\textbf{Logit Adjustment (LA)} enhances model performance on novel categories by mitigating category bias (CBM) and transferring knowledge (CCM). The removal of the weighting mechanism results in a performance decline for known categories due to indiscriminate penalization of their logits. 
\textbf{Training} with the combined loss function is vital for learning both known and novel categories, and initializing the classifier with prototypes offers essential prior knowledge for novel categories.
Discussions about \textbf{real-world applications} will be included in a later section.

\begin{figure}[t]
\centering
\includegraphics[width=0.39\textwidth]{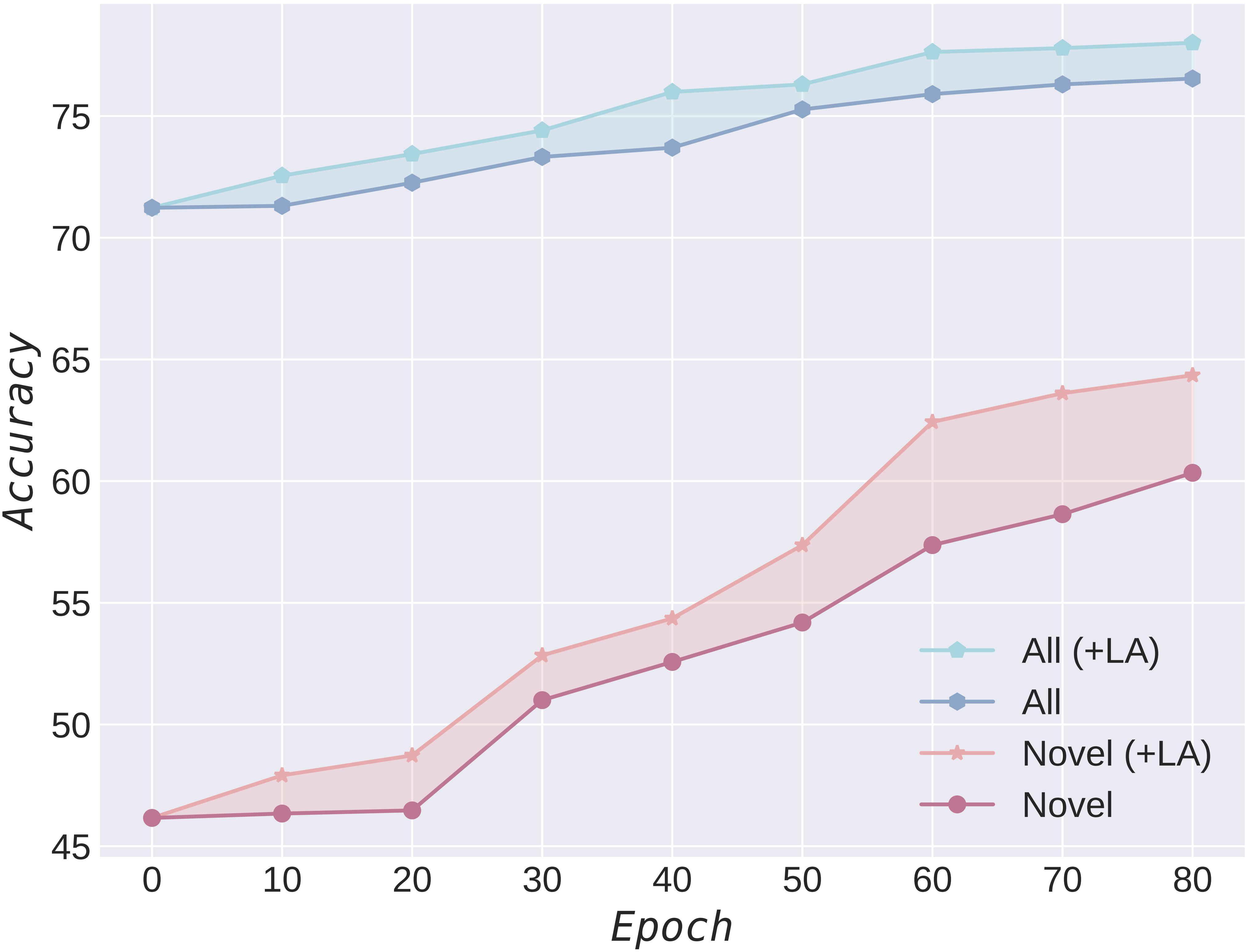}
\captionof{figure}{Accuracy of pseudo labels for all (All) and novel (Novel) categories with and without logit adjustment (LA).}
\label{fig4}
\end{figure}

\subsubsection{Accuracy for Pseudo Labels.}
Given that the proposed logit adjustment strategy aims to generate more accurate pseudo labels for model training, we visualize the accuracy of pseudo labels during training with and without our logit adjustment (LA) strategy in Fig. \ref{fig4}. The results indicate that our model continuously improves accuracy across all categories, with a notable improvement for novel categories. These findings illustrate the effectiveness of our logit adjustment strategy in generating high-quality pseudo labels for novel categories, thereby enhancing model performance in these categories.

\subsubsection{Feature Visualization.}
We utilize t-SNE to visualize the features from 15 categories learned by our model on the CLINC dataset, as depicted in Fig. 5 (a). The results reveal that our model achieves more separable and compact feature distributions after training on the unlabeled data, thereby underscoring our model's effectiveness.

\subsection{Real-world Applications}
In this section, we examine the robustness of our model towards various real-world scenarios.
\subsubsection{Effect of the Known Category Ratio.}
The known category ratio can vary in real-world applications. To investigate its impact on model performance, we adjust this ratio within the set \{0.25, 0.50, 0.75\}. As illustrated in Fig. 5 (b), our model demonstrates superior performance in terms of accuracy for novel categories across different settings, highlighting the model's effectiveness and robustness.

\subsubsection{Number of categories estimation.}
Following previous work \cite{dpn,ktn}, we assume the number of categories $K$ is known in our experiments. However, the number is not always available in real-world applications. Therefore, we address the challenge of estimating $K$ using the dropout algorithm \cite{thu2021}. As shown in Table \ref{table4} Top, our estimations closely approximate the true number of categories, validating the effectiveness of our model.

\subsubsection{Training without Ground-truth $K$.}
In addition to obtaining close estimations of $K$, we also conduct experiments using the estimated $K$. As shown in Table \ref{table2} (w/o Ground-truth $K$), our model achieves comparable results even without the actual $K$, showing the robustness of our approach.

\begin{figure}
\centering
\subfigure[Feature visualization]{
\begin{minipage}[t]{0.45\linewidth}
\centering
\includegraphics[width=1\textwidth]{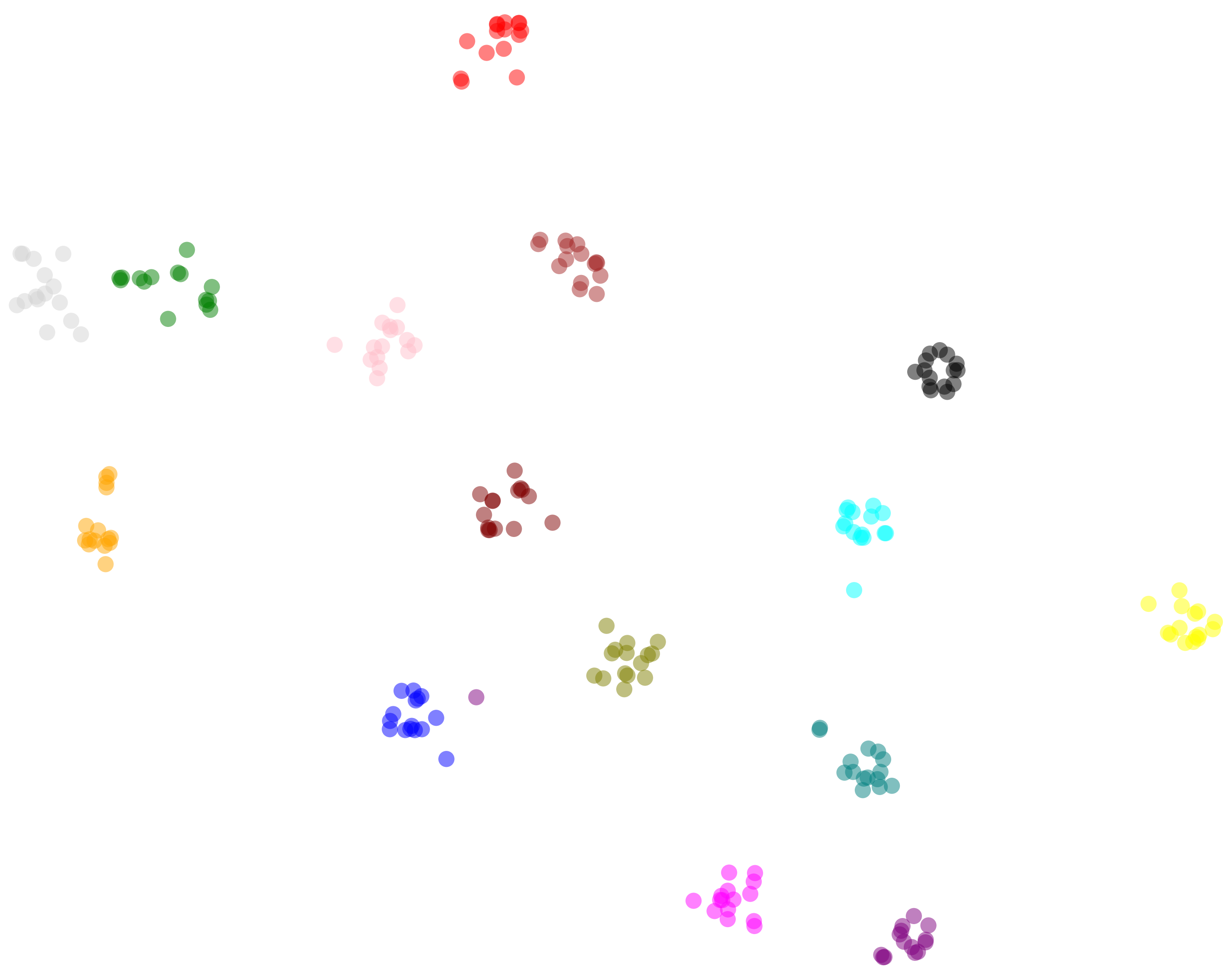}
\end{minipage}%
}%
\subfigure[Accuracy with different KCR]{
\begin{minipage}[t]{0.48\linewidth}
\centering
\includegraphics[width=1\textwidth]{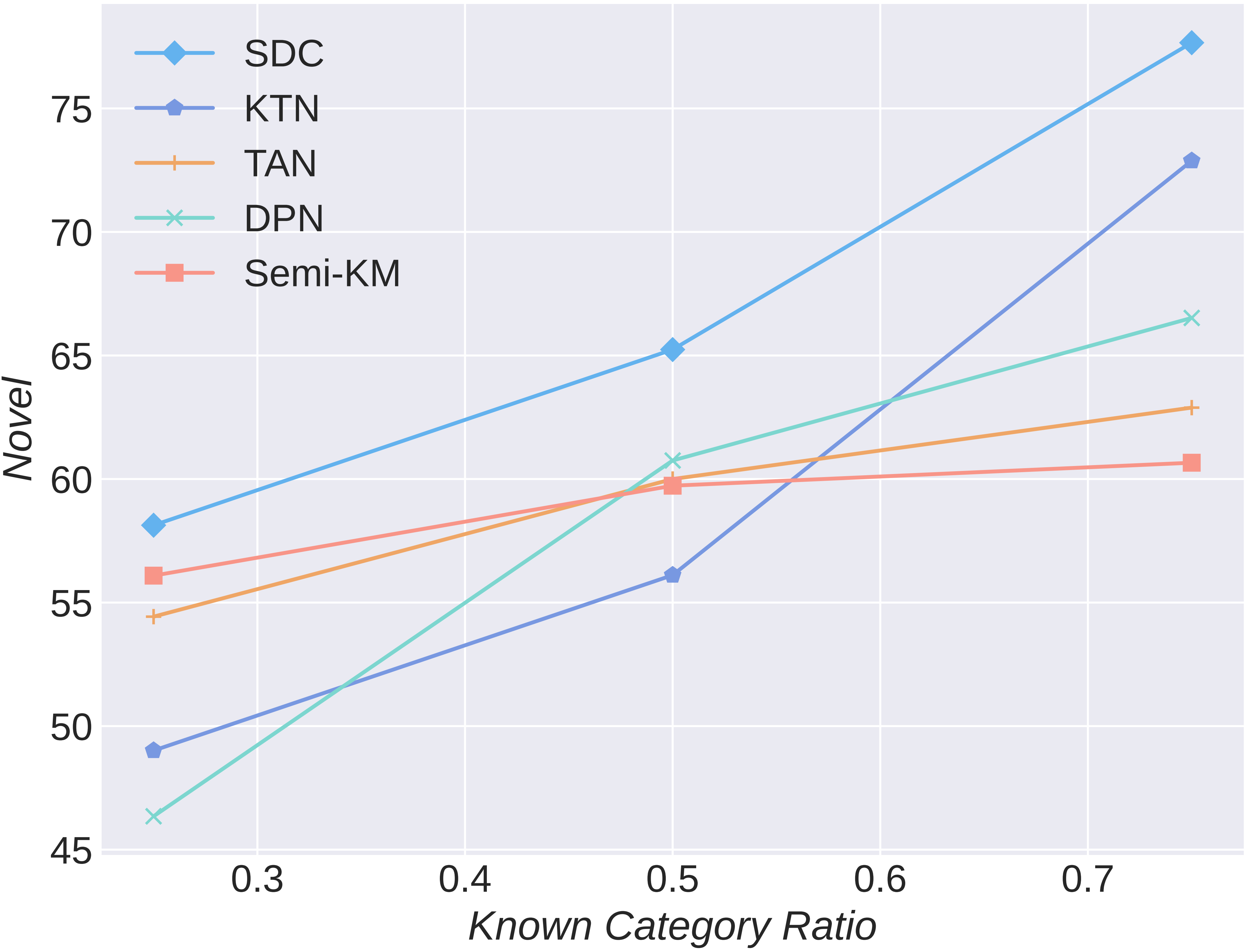}
\end{minipage}%
} 

\centering
\caption{Visualization of the learned features and model performance under different known category ratios (KCR).}
\label{fig5}
\end{figure}

\subsubsection{Online Inference.}
We report experimental results using clustering on the testing sets offline to ensure a fair comparison with previous work. Owing to the learned classifier, our model is also capable of performing online inference for streaming data without the necessity of accumulating abundant data for clustering. As demonstrated in Table \ref{table2} (Online Inference), our model achieves comparable results using the classifier for inference while offering a significantly faster inference speed (average 5.6x faster in Table \ref{table4} Bottom).

\begin{table}
\centering
\begin{tabular}{lccc}
\toprule
Dataset & BANKING  & HWU64  & CLINC \\
\midrule
Ground Truth       & 77    & 64 & 150\\
\rowcolor{gray!20} Estimation & $77.7_{(\pm1.56)}$    & $64_{(\pm2.16)}$ & $148_{(\pm2.82)}$ \\
\midrule
\midrule
Clustering                    & 22.2s    & 6.7s & 27.5s \\
\rowcolor{gray!20} Classifier & 3.3s    & 0.8s & 1.5s \\
\bottomrule
\end{tabular}
\caption{Estimation of $K$ (Top) and inference time (Bottom).}
\label{table4}
\end{table}

\section{Conclusion}
In this paper, we introduce the \textit{Self-Debiasing Calibration} (SDC), a novel framework designed to address the model bias inherent in GCD. Unlike traditional methods that perceive model bias as a hindrance, SDC leverages the bias to improve model performance, particularly in recognizing novel categories. 
Specifically, the core innovation of SDC lies in its dual utilization of the biased output from the pre-trained model.
First, the biased output serves as an accurate and explicit representation of category bias, which can be utilized to debias the output of the current model, thereby producing less biased predictions. Second, the biased output facilitates the differentiation between novel categories by transferring knowledge between similar categories.
This dual approach enables SDC to dynamically adjust the logits of the current model, resulting in less biased logits and more accurate pseudo labels for unlabeled data, effectively addressing the issues of category bias and category confusion. Our experiments on three benchmark datasets demonstrate the superiority of SDC over SOTA methods, particularly in the context of novel category identification.

\section{ACKNOWLEDGMENTS}
This work was supported by the National Science and Technology Major Project (2022ZD0117102), National Natural Science Foundation of China (62293551, 62177038, 62277042, 62137002, 61937001, 62377038). Project of China Knowledge Centre for Engineering Science and Technology, ``LENOVO-XJTU" Intelligent Industry Joint Laboratory Project.

\bibliography{aaai25}

\begin{thebibliography}{45}
\providecommand{\natexlab}[1]{#1}

\bibitem[{Achiam et~al.(2023)Achiam, Adler, Agarwal, Ahmad, Akkaya, Aleman, Almeida, Altenschmidt, Altman, Anadkat et~al.}]{gpt4}
Achiam, J.; Adler, S.; Agarwal, S.; Ahmad, L.; Akkaya, I.; Aleman, F.~L.; Almeida, D.; Altenschmidt, J.; Altman, S.; Anadkat, S.; et~al. 2023.
\newblock Gpt-4 technical report.
\newblock \emph{arXiv preprint arXiv:2303.08774}.

\bibitem[{An et~al.(2024{\natexlab{a}})An, Shi, Tian, Lin, Wang, Wu, Cai, Wang, Chen, Zhu, and Chen}]{loop}
An, W.; Shi, W.; Tian, F.; Lin, H.; Wang, Q.; Wu, Y.; Cai, M.; Wang, L.; Chen, Y.; Zhu, H.; and Chen, P. 2024{\natexlab{a}}.
\newblock Generalized Category Discovery with Large Language Models in the Loop.
\newblock In \emph{Findings of the Association for Computational Linguistics: ACL 2024}, 8653--8665. Bangkok, Thailand: Association for Computational Linguistics.

\bibitem[{An et~al.(2022)An, Tian, Chen, Tang, Zheng, and Wang}]{fcdc}
An, W.; Tian, F.; Chen, P.; Tang, S.; Zheng, Q.; and Wang, Q. 2022.
\newblock Fine-grained Category Discovery under Coarse-grained supervision with Hierarchical Weighted Self-contrastive Learning.
\newblock In \emph{Proceedings of the 2022 Conference on Empirical Methods in Natural Language Processing}, 1314--1323.

\bibitem[{An et~al.(2023{\natexlab{a}})An, Tian, Chen, Zheng, and Ding}]{ptjn}
An, W.; Tian, F.; Chen, P.; Zheng, Q.; and Ding, W. 2023{\natexlab{a}}.
\newblock New User Intent Discovery with Robust Pseudo Label Training and Source Domain Joint-training.
\newblock \emph{IEEE Intelligent Systems}.

\bibitem[{An et~al.(2024{\natexlab{b}})An, Tian, Shi, Chen, Wu, Wang, and Chen}]{tan}
An, W.; Tian, F.; Shi, W.; Chen, Y.; Wu, Y.; Wang, Q.; and Chen, P. 2024{\natexlab{b}}.
\newblock Transfer and alignment network for generalized category discovery.
\newblock In \emph{Proceedings of the AAAI Conference on Artificial Intelligence}, volume~38, 10856--10864.

\bibitem[{An et~al.(2023{\natexlab{b}})An, Tian, Shi, Chen, Zheng, Wang, and Chen}]{dna}
An, W.; Tian, F.; Shi, W.; Chen, Y.; Zheng, Q.; Wang, Q.; and Chen, P. 2023{\natexlab{b}}.
\newblock DNA: Denoised Neighborhood Aggregation for Fine-grained Category Discovery.
\newblock In \emph{Proceedings of the 2023 Conference on Empirical Methods in Natural Language Processing}, 12292--12302.

\bibitem[{An et~al.(2024{\natexlab{c}})An, Tian, Shi, Lin, Wu, Cai, Wang, Wen, Yao, and Chen}]{down}
An, W.; Tian, F.; Shi, W.; Lin, H.; Wu, Y.; Cai, M.; Wang, L.; Wen, H.; Yao, L.; and Chen, P. 2024{\natexlab{c}}.
\newblock Down: Dynamic order weighted network for fine-grained category discovery.
\newblock \emph{Knowledge-Based Systems}, 293: 111666.

\bibitem[{An et~al.(2023{\natexlab{c}})An, Tian, Zheng, Ding, Wang, and Chen}]{dpn}
An, W.; Tian, F.; Zheng, Q.; Ding, W.; Wang, Q.; and Chen, P. 2023{\natexlab{c}}.
\newblock Generalized category discovery with decoupled prototypical network.
\newblock In \emph{Proceedings of the AAAI Conference on Artificial Intelligence}, volume~37, 12527--12535.

\bibitem[{Asano, Rupprecht, and Vedaldi(2019)}]{pseudo}
Asano, Y.~M.; Rupprecht, C.; and Vedaldi, A. 2019.
\newblock Self-labelling via simultaneous clustering and representation learning.
\newblock \emph{arXiv preprint arXiv:1911.05371}.

\bibitem[{Brown et~al.(2020)Brown, Mann, Ryder, Subbiah, Kaplan, Dhariwal, Neelakantan, Shyam, Sastry, Askell et~al.}]{gpt3}
Brown, T.; Mann, B.; Ryder, N.; Subbiah, M.; Kaplan, J.~D.; Dhariwal, P.; Neelakantan, A.; Shyam, P.; Sastry, G.; Askell, A.; et~al. 2020.
\newblock Language models are few-shot learners.
\newblock \emph{Advances in neural information processing systems}, 33: 1877--1901.

\bibitem[{Caron et~al.(2018)Caron, Bojanowski, Joulin, and Douze}]{deepcluster}
Caron, M.; Bojanowski, P.; Joulin, A.; and Douze, M. 2018.
\newblock Deep clustering for unsupervised learning of visual features.
\newblock In \emph{Proceedings of the European Conference on Computer Vision (ECCV)}, 132--149.

\bibitem[{Caron et~al.(2020)Caron, Misra, Mairal, Goyal, Bojanowski, and Joulin}]{sk2}
Caron, M.; Misra, I.; Mairal, J.; Goyal, P.; Bojanowski, P.; and Joulin, A. 2020.
\newblock Unsupervised learning of visual features by contrasting cluster assignments.
\newblock \emph{Advances in neural information processing systems}, 33: 9912--9924.

\bibitem[{Casanueva et~al.(2020)Casanueva, Tem{\v{c}}inas, Gerz, Henderson, and Vuli{\'c}}]{banking}
Casanueva, I.; Tem{\v{c}}inas, T.; Gerz, D.; Henderson, M.; and Vuli{\'c}, I. 2020.
\newblock Efficient intent detection with dual sentence encoders.
\newblock \emph{arXiv preprint arXiv:2003.04807}.

\bibitem[{Chen et~al.(2022)Chen, Shen, Zhang, and Torr}]{zero1}
Chen, D.; Shen, Y.; Zhang, H.; and Torr, P.~H. 2022.
\newblock Zero-shot logit adjustment.
\newblock \emph{arXiv preprint arXiv:2204.11822}.

\bibitem[{Cui et~al.(2019)Cui, Jia, Lin, Song, and Belongie}]{tail4}
Cui, Y.; Jia, M.; Lin, T.-Y.; Song, Y.; and Belongie, S. 2019.
\newblock Class-balanced loss based on effective number of samples.
\newblock In \emph{Proceedings of the IEEE/CVF conference on computer vision and pattern recognition}, 9268--9277.

\bibitem[{Cuturi(2013)}]{sk}
Cuturi, M. 2013.
\newblock Sinkhorn distances: Lightspeed computation of optimal transport.
\newblock \emph{Advances in neural information processing systems}, 26.

\bibitem[{Devlin et~al.(2018)Devlin, Chang, Lee, and Toutanova}]{bert}
Devlin, J.; Chang, M.-W.; Lee, K.; and Toutanova, K. 2018.
\newblock Bert: Pre-training of deep bidirectional transformers for language understanding.
\newblock \emph{arXiv preprint arXiv:1810.04805}.

\bibitem[{Gidaris and Komodakis(2018)}]{zero2}
Gidaris, S.; and Komodakis, N. 2018.
\newblock Dynamic few-shot visual learning without forgetting.
\newblock In \emph{Proceedings of the IEEE conference on computer vision and pattern recognition}, 4367--4375.

\bibitem[{Han, Vedaldi, and Zisserman(2019)}]{dtc}
Han, K.; Vedaldi, A.; and Zisserman, A. 2019.
\newblock Learning to discover novel visual categories via deep transfer clustering.
\newblock In \emph{Proceedings of the IEEE/CVF International Conference on Computer Vision}, 8401--8409.

\bibitem[{Larson et~al.(2019)Larson, Mahendran, Peper, Clarke, Lee, Hill, Kummerfeld, Leach, Laurenzano, Tang et~al.}]{clinc}
Larson, S.; Mahendran, A.; Peper, J.~J.; Clarke, C.; Lee, A.; Hill, P.; Kummerfeld, J.~K.; Leach, K.; Laurenzano, M.~A.; Tang, L.; et~al. 2019.
\newblock An evaluation dataset for intent classification and out-of-scope prediction.
\newblock \emph{arXiv preprint arXiv:1909.02027}.

\bibitem[{Lin et~al.(2024{\natexlab{a}})Lin, An, Chen, Tian, Yao, Ding, Wang, and Chen}]{tribranch}
Lin, H.; An, W.; Chen, Y.; Tian, F.; Yao, Y.; Ding, W.; Wang, Q.; and Chen, P. 2024{\natexlab{a}}.
\newblock A Tri-Branch Network with Prototype-aware Matching for Universal Category Discovery.
\newblock In \emph{2024 IEEE International Conference on Multimedia and Expo (ICME)}, 1--6. IEEE Computer Society.

\bibitem[{Lin et~al.(2024{\natexlab{b}})Lin, An, Wang, Chen, Tian, Wang, Wang, Dai, and Wang}]{flip}
Lin, H.; An, W.; Wang, J.; Chen, Y.; Tian, F.; Wang, M.; Wang, Q.; Dai, G.; and Wang, J. 2024{\natexlab{b}}.
\newblock Flipped Classroom: Aligning Teacher Attention with Student in Generalized Category Discovery.
\newblock In \emph{The Thirty-eighth Annual Conference on Neural Information Processing Systems}.

\bibitem[{Lin, Xu, and Zhang(2020)}]{thu2020}
Lin, T.-E.; Xu, H.; and Zhang, H. 2020.
\newblock Discovering new intents via constrained deep adaptive clustering with cluster refinement.
\newblock In \emph{Proceedings of the AAAI Conference on Artificial Intelligence}, volume~34, 8360--8367.

\bibitem[{Liu et~al.(2021)Liu, Eshghi, Swietojanski, and Rieser}]{hwu64}
Liu, X.; Eshghi, A.; Swietojanski, P.; and Rieser, V. 2021.
\newblock Benchmarking natural language understanding services for building conversational agents.
\newblock In \emph{Increasing Naturalness and Flexibility in Spoken Dialogue Interaction}. Springer.

\bibitem[{MacQueen et~al.(1967)}]{km}
MacQueen, J.; et~al. 1967.
\newblock Some methods for classification and analysis of multivariate observations.
\newblock In \emph{Proceedings of the fifth Berkeley symposium on mathematical statistics and probability}, volume~1, 281--297. Oakland, CA, USA.

\bibitem[{Menon et~al.(2020)Menon, Jayasumana, Rawat, Jain, Veit, and Kumar}]{tail1}
Menon, A.~K.; Jayasumana, S.; Rawat, A.~S.; Jain, H.; Veit, A.; and Kumar, S. 2020.
\newblock Long-tail learning via logit adjustment.
\newblock \emph{arXiv preprint arXiv:2007.07314}.

\bibitem[{Pennington, Socher, and Manning(2014)}]{glove}
Pennington, J.; Socher, R.; and Manning, C.~D. 2014.
\newblock Glove: Global vectors for word representation.
\newblock In \emph{Proceedings of the 2014 conference on empirical methods in natural language processing (EMNLP)}, 1532--1543.

\bibitem[{Rastegar, Doughty, and Snoek(2024)}]{rastegar2024learn}
Rastegar, S.; Doughty, H.; and Snoek, C. 2024.
\newblock Learn to categorize or categorize to learn? self-coding for generalized category discovery.
\newblock \emph{Advances in Neural Information Processing Systems}, 36.

\bibitem[{Saito and Saenko(2021)}]{hscore}
Saito, K.; and Saenko, K. 2021.
\newblock Ovanet: One-vs-all network for universal domain adaptation.
\newblock In \emph{Proceedings of the ieee/cvf international conference on computer vision}, 9000--9009.

\bibitem[{Shi et~al.(2024)Shi, An, Tian, Chen, Wu, Wang, and Chen}]{ktn}
Shi, W.; An, W.; Tian, F.; Chen, Y.; Wu, Y.; Wang, Q.; and Chen, P. 2024.
\newblock A Unified Knowledge Transfer Network for Generalized Category Discovery.
\newblock In \emph{Proceedings of the AAAI Conference on Artificial Intelligence}, volume~38, 18961--18969.

\bibitem[{Shi et~al.(2023)Shi, An, Tian, Zheng, Wang, and Chen}]{diffusion}
Shi, W.; An, W.; Tian, F.; Zheng, Q.; Wang, Q.; and Chen, P. 2023.
\newblock A Diffusion Weighted Graph Framework for New Intent Discovery.
\newblock In \emph{Proceedings of the 2023 Conference on Empirical Methods in Natural Language Processing}, 8033--8042.

\bibitem[{Vaze et~al.(2022)Vaze, Han, Vedaldi, and Zisserman}]{gcd}
Vaze, S.; Han, K.; Vedaldi, A.; and Zisserman, A. 2022.
\newblock Generalized category discovery.
\newblock In \emph{Proceedings of the IEEE/CVF Conference on Computer Vision and Pattern Recognition}, 7492--7501.

\bibitem[{Vaze, Vedaldi, and Zisserman(2024)}]{vaze2024no}
Vaze, S.; Vedaldi, A.; and Zisserman, A. 2024.
\newblock No representation rules them all in category discovery.
\newblock \emph{Advances in Neural Information Processing Systems}, 36.

\bibitem[{Wen, Zhao, and Qi(2022)}]{simple}
Wen, X.; Zhao, B.; and Qi, X. 2022.
\newblock A Simple Parametric Classification Baseline for Generalized Category Discovery.
\newblock \emph{arXiv preprint arXiv:2211.11727}.

\bibitem[{Wolf et~al.(2019)Wolf, Debut, Sanh, Chaumond, Delangue, Moi, Cistac, Rault, Louf, Funtowicz et~al.}]{huggingface}
Wolf, T.; Debut, L.; Sanh, V.; Chaumond, J.; Delangue, C.; Moi, A.; Cistac, P.; Rault, T.; Louf, R.; Funtowicz, M.; et~al. 2019.
\newblock HuggingFace's Transformers: State-of-the-art natural language processing.
\newblock \emph{arXiv preprint arXiv:1910.03771}.

\bibitem[{Xie, Girshick, and Farhadi(2016)}]{dec}
Xie, J.; Girshick, R.; and Farhadi, A. 2016.
\newblock Unsupervised deep embedding for clustering analysis.
\newblock In \emph{International conference on machine learning}, 478--487. PMLR.

\bibitem[{Yang et~al.(2017)Yang, Fu, Sidiropoulos, and Hong}]{dcn}
Yang, B.; Fu, X.; Sidiropoulos, N.~D.; and Hong, M. 2017.
\newblock Towards k-means-friendly spaces: Simultaneous deep learning and clustering.
\newblock In \emph{international conference on machine learning}, 3861--3870. PMLR.

\bibitem[{Zhang et~al.(2021{\natexlab{a}})Zhang, Xu, Lin, and Lyu}]{thu2021}
Zhang, H.; Xu, H.; Lin, T.-E.; and Lyu, R. 2021{\natexlab{a}}.
\newblock Discovering New Intents with Deep Aligned Clustering.
\newblock In \emph{Proceedings of the AAAI Conference on Artificial Intelligence}.

\bibitem[{Zhang et~al.(2021{\natexlab{b}})Zhang, Bui, Yoon, Chen, Liu, Xia, Tran, Chang, and Yu}]{pretrain}
Zhang, J.; Bui, T.; Yoon, S.; Chen, X.; Liu, Z.; Xia, C.; Tran, Q.~H.; Chang, W.; and Yu, P. 2021{\natexlab{b}}.
\newblock Few-shot intent detection via contrastive pre-training and fine-tuning.
\newblock \emph{arXiv preprint arXiv:2109.06349}.

\bibitem[{Zhang et~al.(2024{\natexlab{a}})Zhang, Yan, Yang, Ren, Bai, Li, and Li}]{zhang2024ronid}
Zhang, S.; Yan, C.; Yang, J.; Ren, C.; Bai, J.; Li, T.; and Li, Z. 2024{\natexlab{a}}.
\newblock RoNID: New Intent Discovery with Generated-Reliable Labels and Cluster-friendly Representations.
\newblock \emph{arXiv preprint arXiv:2404.08977}.

\bibitem[{Zhang et~al.(2024{\natexlab{b}})Zhang, Yang, Bai, Yan, Li, Yan, and Li}]{zhang2024new}
Zhang, S.; Yang, J.; Bai, J.; Yan, C.; Li, T.; Yan, Z.; and Li, Z. 2024{\natexlab{b}}.
\newblock New Intent Discovery with Attracting and Dispersing Prototype.
\newblock \emph{arXiv preprint arXiv:2403.16913}.

\bibitem[{Zhang et~al.(2022)Zhang, Zhang, Zhan, Wu, and Lam}]{mtp}
Zhang, Y.; Zhang, H.; Zhan, L.-M.; Wu, X.-M.; and Lam, A. 2022.
\newblock New intent discovery with pre-training and contrastive learning.
\newblock \emph{arXiv preprint arXiv:2205.12914}.

\bibitem[{Zhao et~al.(2022)Zhao, Chen, Tan, Huang, and Zhu}]{tail2}
Zhao, Y.; Chen, W.; Tan, X.; Huang, K.; and Zhu, J. 2022.
\newblock Adaptive logit adjustment loss for long-tailed visual recognition.
\newblock In \emph{Proceedings of the AAAI conference on artificial intelligence}, volume~36, 3472--3480.

\bibitem[{Zhao et~al.(2024)Zhao, Li, Zhai, and Chang}]{zhao2024pseudo}
Zhao, Z.; Li, X.; Zhai, Z.; and Chang, Z. 2024.
\newblock Pseudo-supervised contrastive learning with inter-class separability for generalized category discovery.
\newblock \emph{Knowledge-Based Systems}, 289: 111477.

\bibitem[{Zhong et~al.(2021)Zhong, Fini, Roy, Luo, Ricci, and Sebe}]{ncl}
Zhong, Z.; Fini, E.; Roy, S.; Luo, Z.; Ricci, E.; and Sebe, N. 2021.
\newblock Neighborhood contrastive learning for novel class discovery.
\newblock In \emph{Proceedings of the IEEE/CVF conference on computer vision and pattern recognition}, 10867--10875.

\end{thebibliography}

\end{document}